\documentclass[letterpaper,journal]{IEEEtran}

\usepackage{amsmath,amsfonts,amssymb}
\usepackage{array}
\usepackage{textcomp}
\usepackage{stfloats}
\usepackage{url}
\usepackage{graphicx}
\usepackage{cite}
\usepackage{multirow}
\usepackage{booktabs}
\usepackage{color}
\usepackage[hidelinks]{hyperref}

\begin{document}

\title{Counterfactual Motion Reliability Learning for Robust UAV Tracking}

\author{
Yuehai Chen,
Jian Lan,~\IEEEmembership{Senior Member,~IEEE},
and Yuan Wei%
\thanks{This work was supported in part by the National Natural Science Foundation of China under Grants U23B2035 and 62273269.
\emph{(Corresponding author: Jian Lan.)}}%
\thanks{Yuehai Chen and Yuan Wei are with the Faculty of Electronics and Information Engineering and the Institute for Low-Altitude Regulation, Xi'an Jiaotong University, Xi'an 710049, China
(e-mail: \href{mailto:chenyuehai@xjtu.edu.cn}{chenyuehai@xjtu.edu.cn};
\href{mailto:weiyuan98@stu.xjtu.edu.cn}{weiyuan98@stu.xjtu.edu.cn}).}%
\thanks{Jian Lan is with the Institute for Low-Altitude Regulation, Xi'an Jiaotong University, Xi'an 710049, China, and also with Lanzhou University, Lanzhou 730000, China
(e-mail: \href{mailto:lanjian@mail.xjtu.edu.cn}{lanjian@mail.xjtu.edu.cn}).}%
}

\maketitle

\begin{abstract}
Infrared unmanned aerial vehicle (UAV) tracking is challenging because the target is often small, low-contrast, and easily confused with thermal distractors or cluttered backgrounds. Recent Transformer-based trackers have achieved promising performance by learning strong appearance representations, but their responses can still be dominated by background structures when the target appearance is weak or ambiguous. A natural solution is to introduce temporal motion cues. However, in infrared UAV tracking, motion cues are not always reliable: camera jitter, dynamic backgrounds, sensor noise, and target disappearance may produce temporal variations that are stronger than the true target motion. Therefore, the key challenge is not simply how to use motion, but how to distinguish target-consistent motion from background-induced pseudo motion. To this end, we propose CMRTrack, a counterfactual motion reliability learning framework for robust infrared UAV tracking. CMRTrack first extracts temporal evidence from adjacent search regions using a lightweight motion evidence encoder. During training, a counterfactual target-erased history branch is introduced to construct hard motion references, encouraging the motion encoder to learn reliable target-consistent motion rather than arbitrary temporal changes. The learned motion evidence is then incorporated into a one-stream tracking framework through motion-guided token modulation and reliability-aware score fusion, enabling adaptive feature enhancement and response refinement. Extensive experiments on Anti-UAV410 demonstrate that CMRTrack consistently outperforms representative state-of-the-art trackers and significantly improves the OSTrack baseline, with ablation studies and qualitative analysis verifying the effectiveness of the proposed counterfactual motion reliability learning.
\end{abstract}


\section{INTRODUCTION}
V{\scshape isual} object tracking aims to localize an arbitrary target in a video sequence given only its initial state, and has been widely studied as a fundamental problem in computer vision. With the rapid proliferation of unmanned aerial vehicles (UAVs), Anti-UAV tracking has become increasingly important for low-altitude security, wide-area surveillance, and counter-UAV systems \cite{TAES3, TAES4, TAES5}. Compared with visible-light imaging, thermal infrared sensing is more suitable for long-range and low-illumination monitoring, making infrared UAV tracking an important yet challenging task. Recent Siamese and Transformer-based trackers have achieved remarkable progress on general tracking benchmarks by learning strong appearance representations and template-search interactions\cite{TAES1, TAES2}. However, directly applying these trackers to infrared UAV scenarios remains difficult due to small target size, weak appearance, thermal distractors, and dynamic backgrounds.

\begin{figure}[t]
    \centering
    \includegraphics[width=0.45\textwidth]{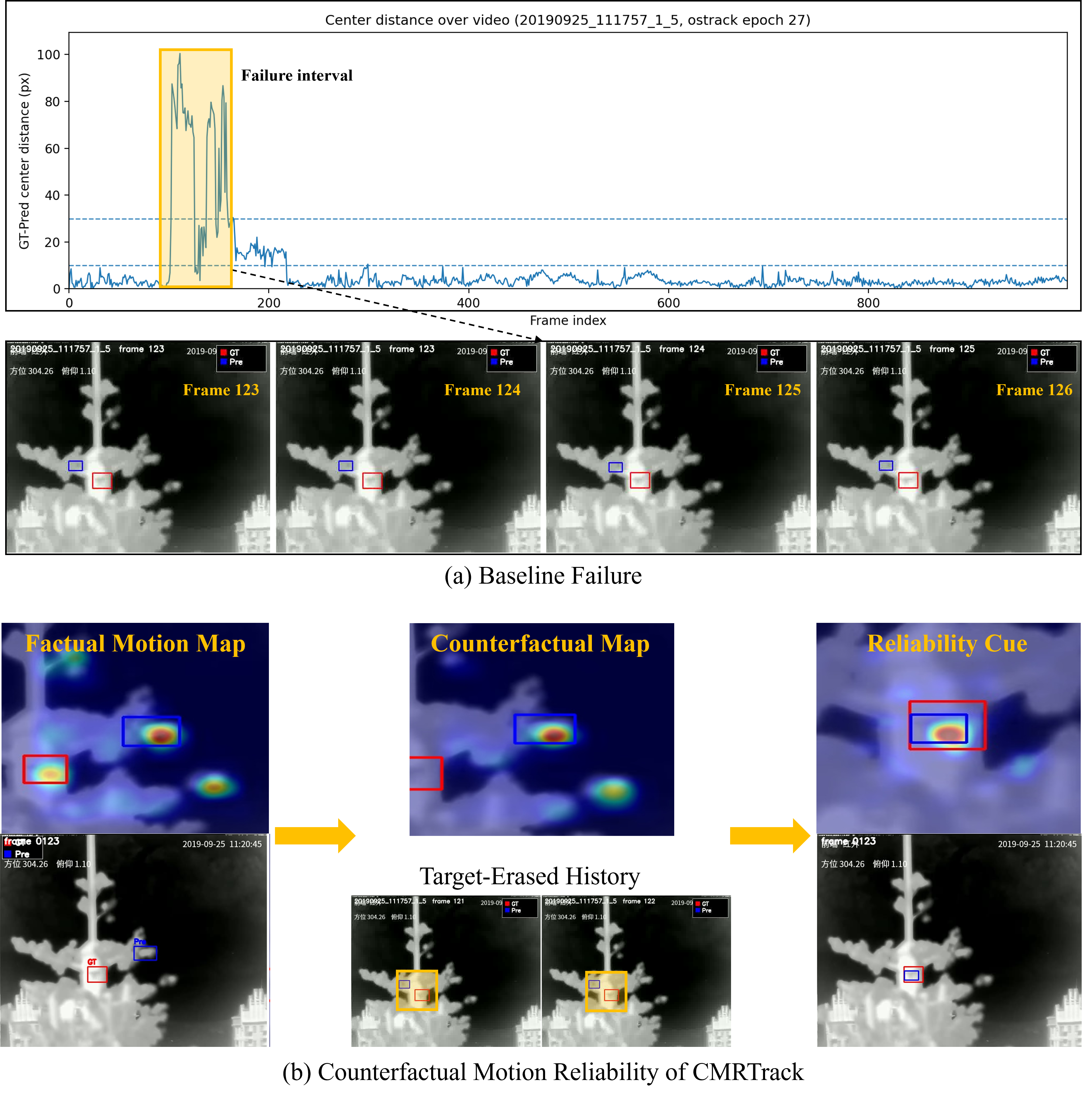}
    \caption{
    Motivation of counterfactual motion reliability learning. The baseline tracker~\cite{ye2022ostrack} drifts to a background distractor in challenging infrared scenes.
    By comparing factual and target-erased counterfactual motion maps, the proposed method suppresses pseudo motion and highlights target-induced motion.
    }
    \label{fig:motivation}
\end{figure}

Despite the strong representation ability of modern trackers, infrared UAV tracking is highly vulnerable to appearance ambiguity. The UAV target usually occupies only a few pixels and lacks discriminative texture, while background structures such as building edges, clouds, and thermal clutter may exhibit target-like intensity patterns. As a result, an appearance-dominant tracker~\cite{ye2022ostrack} may assign high confidence to distractor regions when the target response is weak. As illustrated in Fig.~\ref{fig:motivation}, the baseline tracker is attracted to a background region during the highlighted failure interval, leading to severe localization drift. This observation indicates that appearance matching alone is insufficient for robust infrared UAV tracking, especially when the target is tiny, low-contrast, or surrounded by cluttered backgrounds.

A natural way to alleviate appearance ambiguity is to exploit temporal motion information \cite{10596986}. Motion cues can provide complementary evidence when the UAV target is visually weak or occupies only a small image region. However, in infrared UAV tracking, temporal variations are not necessarily caused by the target. Camera jitter, view adjustment, dynamic backgrounds, sensor noise, and thermal distractors may produce apparent changes that are even stronger than the true target motion. Consequently, a raw motion map may activate both the target region and background structures, introducing background-induced pseudo motion. Therefore, the crucial problem is not simply how to introduce motion cues into the tracker, but how to determine whether a motion response is truly target-consistent.

This motivates us to consider motion reliability from a counterfactual perspective. If the historical target evidence is removed while the current observation is kept unchanged, motion responses that truly depend on the target trajectory should become less consistent or exhibit a clear response gap, whereas responses caused by background displacement, camera motion, or sensor noise may still remain. Such a counterfactual comparison provides a useful reference for identifying target-induced motion. Based on this observation, we construct a target-erased historical search region during training and compare the factual motion response with its counterfactual counterpart. The response gap between them is used to encourage the tracker to emphasize reliable target-consistent motion and suppress background-induced pseudo motion.

Based on this insight, we propose CMRTrack, a counterfactual motion reliability learning framework built upon a one-stream Transformer tracker. CMRTrack introduces a lightweight motion evidence encoder to estimate temporal motion responses from adjacent search regions. During training, a counterfactual target-erased history branch is constructed to regularize the motion encoder with hard motion references, enabling it to distinguish target-consistent motion from unreliable background variations. The resulting motion evidence is incorporated into the tracking pipeline through motion-guided token modulation and reliability-aware score fusion, enhancing target-aware search representations and refining the final response map. During inference, the counterfactual branch is discarded, so CMRTrack only relies on the factual historical search region without an additional counterfactual forward pass.

The main contributions of this work are summarized as follows:
\begin{itemize}
    \item We propose CMRTrack, a counterfactual motion reliability learning framework for robust infrared UAV tracking. The proposed framework addresses unreliable motion responses in infrared UAV scenarios by explicitly distinguishing target-consistent motion from background-induced pseudo motion.

    \item We design a counterfactual target-erased history strategy for motion reliability learning. By comparing factual and counterfactual motion responses, CMRTrack constructs hard references for unreliable motion and imposes relative constraints that encourage the motion encoder to preserve target-induced motion evidence while suppressing background responses.

    \item We integrate reliable motion evidence into a one-stream tracking framework through motion-guided token modulation and reliability-aware score fusion. Extensive experiments on Anti-UAV410, including state-of-the-art comparison, ablation study, attribute-based evaluation, and qualitative analysis, demonstrate the effectiveness and robustness of the proposed method.
\end{itemize}

\section{Related Work}

\subsection{General Visual Object Tracking}

Visual object tracking aims to localize an arbitrary target in subsequent frames given its initial state. Deep trackers have achieved substantial progress by learning discriminative target representations and effective template-search matching functions. SiamFC~\cite{bertinetto2016siamfc} introduces a fully-convolutional Siamese framework for efficient similarity matching, while SiamRPN~\cite{li2018siamrpn} and SiamRPN++~\cite{li2019siamrpnpp} further improve Siamese tracking with region proposal prediction and stronger backbone networks. In parallel, discriminative trackers such as ATOM~\cite{danelljan2019atom} and DiMP~\cite{bhat2019dimp} learn target-specific prediction models to improve localization accuracy and robustness.

Transformer-based trackers have recently become a dominant direction due to their strong feature interaction ability. TransT~\cite{chen2021transt} introduces Transformer attention to model the relation between template and search features, and STARK~\cite{yan2021stark} exploits spatio-temporal Transformer representations for robust tracking. MixFormer~\cite{cui2022mixformer} designs an end-to-end mixed-attention framework to jointly model target appearance and localization. Different from these two-stream or multi-stage designs, OSTrack~\cite{ye2022ostrack} proposes a one-stream framework that directly concatenates template and search tokens, unifying feature extraction and relation modeling in a single Vision Transformer. Our method adopts this efficient one-stream tracking paradigm as the baseline, but extends it with counterfactual motion reliability learning to improve robustness in infrared UAV tracking.

\subsection{Anti-UAV Tracking}

Anti-UAV tracking is more challenging than general visual object tracking because UAV targets are often tiny, low-contrast, and frequently affected by thermal infrared noise, background clutter, abrupt camera motion, and target disappearance. The Anti-UAV benchmark~\cite{jiang2023antiuav} and Anti-UAV410~\cite{huang2024antiuav410} provide important datasets and evaluation protocols for this task, where trackers are required to estimate both target localization and target existence state. Compared with general tracking datasets such as GOT-10k~\cite{huang2019got10k}, LaSOT~\cite{fan2021lasot}, and TrackingNet~\cite{muller2018trackingnet}, Anti-UAV tracking places higher demands on robustness to long-range imaging degradation, infrared distractors, dynamic backgrounds, and out-of-view cases.

Existing Anti-UAV and UAV tracking methods improve robustness from different perspectives. GASiam~\cite{shi2022gasiam} enhances infrared Anti-UAV feature representation with graph attention. Global search and re-detection methods, such as GlobalTrack~\cite{huang2020globaltrack} and Siam R-CNN~\cite{voigtlaender2020siamrcnn}, improve recovery ability when the target is lost, but usually introduce higher computational cost. FocusTrack~\cite{wang2025focustrack} bridges local and global tracking by adaptively adjusting the search region according to target presence probability. Different from these methods, our work does not mainly focus on enlarging the search region or performing global re-detection. Instead, we investigate whether the motion response used by the tracker is truly induced by the UAV target, which is critical for suppressing background-induced pseudo motion in infrared scenes.

\subsection{Motion and Temporal Cues}

Motion and temporal cues are important for robust tracking, especially when the target appearance is weak or ambiguous. Several trackers exploit temporal information to improve target localization. STARK~\cite{yan2021stark} introduces spatio-temporal Transformer representations for visual tracking, while TCTrack~\cite{cao2022tctrack} explicitly uses temporal contexts for aerial tracking. Autoregressive trackers further model temporal dependencies by predicting tracking results in a sequential manner~\cite{wei2023artrack,xie2024aqatrack}. These methods demonstrate that temporal modeling can improve tracking robustness under appearance variation and motion uncertainty.

Motion cues are particularly useful for tiny UAV and infrared target perception, where the target often occupies only a few pixels and lacks discriminative texture. Video tiny-object detection guided by spatial-temporal motion information~\cite{yang2023video} shows that temporal motion can improve tiny object detection in videos. Bio-inspired magnocellular computation~\cite{wang2024tiny} also demonstrates that motion-sensitive mechanisms can enhance tiny drone perception. More recently, MCATrack~\cite{zhang2025mcatrack} introduces a motion-centric adaptive tracking framework for infrared tiny drone tracking by enhancing local motion-sensitive responses against cluttered backgrounds.

Although motion information is beneficial, directly relying on motion responses can be risky in infrared UAV scenarios. Camera jitter, moving backgrounds, thermal distractors, and sensor noise may produce target-like temporal variations, causing motion-aware trackers to focus on unreliable regions. Existing temporal or motion-aware trackers mainly emphasize how to extract, enhance, or aggregate motion cues, but they rarely verify whether the observed motion is truly caused by the target. In contrast, our method explicitly models motion reliability from a counterfactual perspective. By erasing the historical target region and comparing factual and counterfactual motion responses, CMRTrack learns to distinguish target-consistent motion from background-induced pseudo motion.

\section{Proposed Method}

\subsection{Overview}

\begin{figure*}[t]
    \centering
    \includegraphics[width=\textwidth]{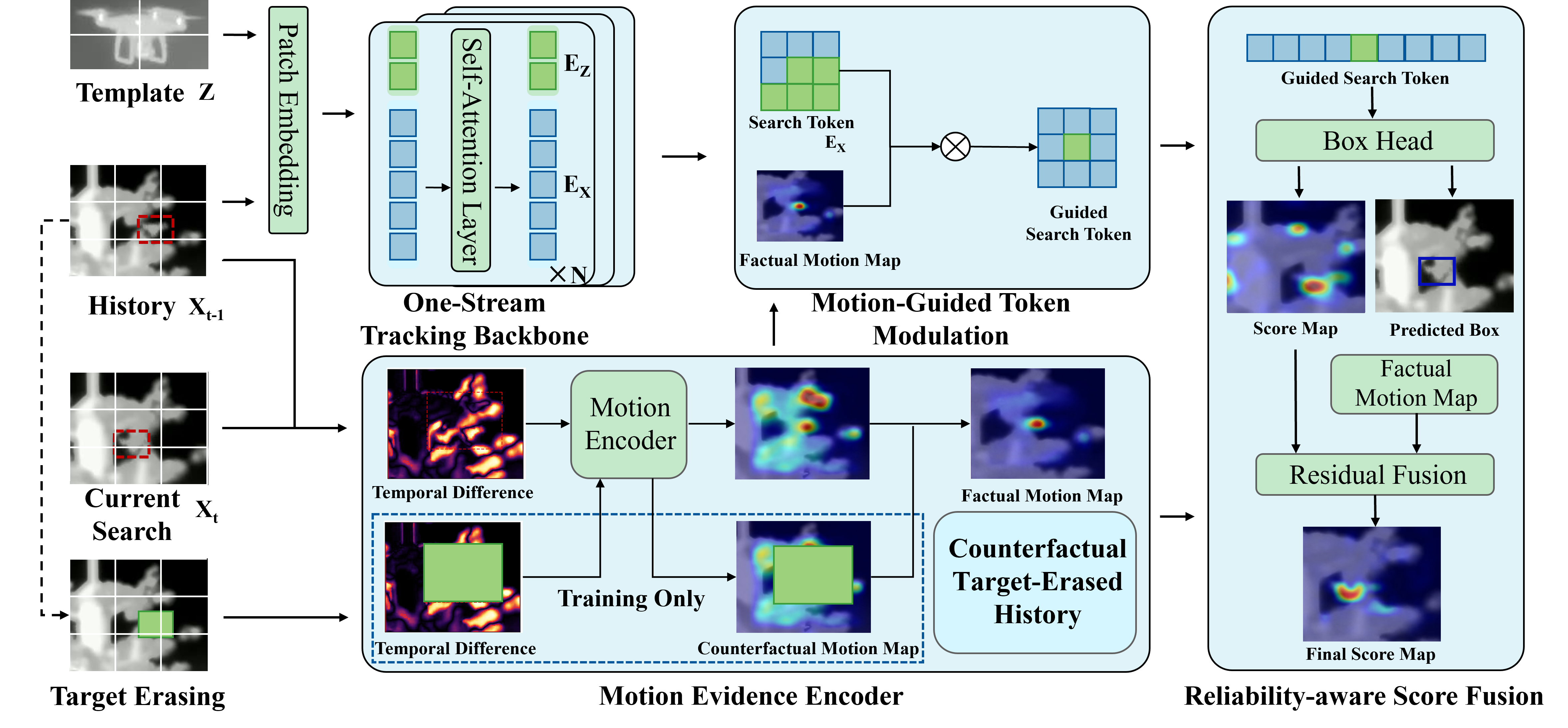}
    \caption{
    Overall framework of the proposed CMRTrack. The factual motion map $M_t$ is generated from the temporal difference between the current search region $X_t$ and the historical search region $X_{t-1}$, and is used for both motion-guided token modulation and reliability-aware score fusion. During training, a counterfactual target-erased history $\tilde{X}_{t-1}$ is constructed to generate the counterfactual motion map $\tilde{M}_t$, which provides reliability constraints for suppressing background-induced pseudo motion. The counterfactual branch is removed during inference.
    }
    \label{fig:framework}
\end{figure*}

We propose CMRTrack, a counterfactual motion reliability learning framework for infrared UAV tracking. As illustrated in Fig.~\ref{fig:framework}, CMRTrack is built upon a one-stream Transformer tracker. Given the template image $Z$, the current search region $X_t$, and the historical search region $X_{t-1}$, the one-stream backbone extracts target-aware search tokens through joint template-search interaction. In parallel, a lightweight motion evidence encoder $\Phi$ estimates a factual motion map $M_t$ from the temporal difference between $X_t$ and $X_{t-1}$. The factual motion map is then used to guide the tracking process in two ways: it modulates search tokens before localization and refines the appearance score map through reliability-aware score fusion.

During training, we further construct a counterfactual target-erased history $\tilde{X}_{t-1}$ by removing the historical target region from $X_{t-1}$. The same motion evidence encoder is applied to $(X_t,\tilde{X}_{t-1})$ to generate a counterfactual motion map $\tilde{M}_t$. By comparing the factual and counterfactual motion responses, CMRTrack learns to emphasize target-consistent motion while suppressing background-induced pseudo motion. The counterfactual branch is used only for training-time reliability learning. During inference, CMRTrack operates causally with the normal historical search region, and no counterfactual forward pass is required.

\subsection{One-Stream Tracking Backbone}

Given a template image $Z$ and the current search region $X_t$, CMRTrack follows the one-stream tracking paradigm to perform joint template-search feature interaction. Specifically, $Z$ and $X_t$ are first divided into patch tokens and then concatenated as the input of a shared Vision Transformer backbone. Through stacked self-attention layers, the template tokens provide target-specific guidance to the search tokens, producing target-aware search features. Let $\mathcal{B}(\cdot)$ denote the one-stream backbone and $\mathcal{H}(\cdot)$ denote the localization head. The standard appearance-based tracking output can be written as
\begin{equation}
    \{B_t, S_t\} = \mathcal{H}(\mathcal{B}(Z, X_t)),
\end{equation}
where $B_t$ denotes the predicted bounding box and $S_t$ denotes the appearance score map.

This backbone provides the basic localization ability of CMRTrack. However, in infrared UAV tracking, the target is often tiny, low-contrast, or visually similar to the background, making the appearance response alone unreliable. Therefore, we introduce temporal motion evidence as a complementary cue to enhance search features and refine the final response map.

\subsection{Motion Evidence Encoder}

\begin{figure}[t]
    \centering
    \includegraphics[width=0.65\linewidth]{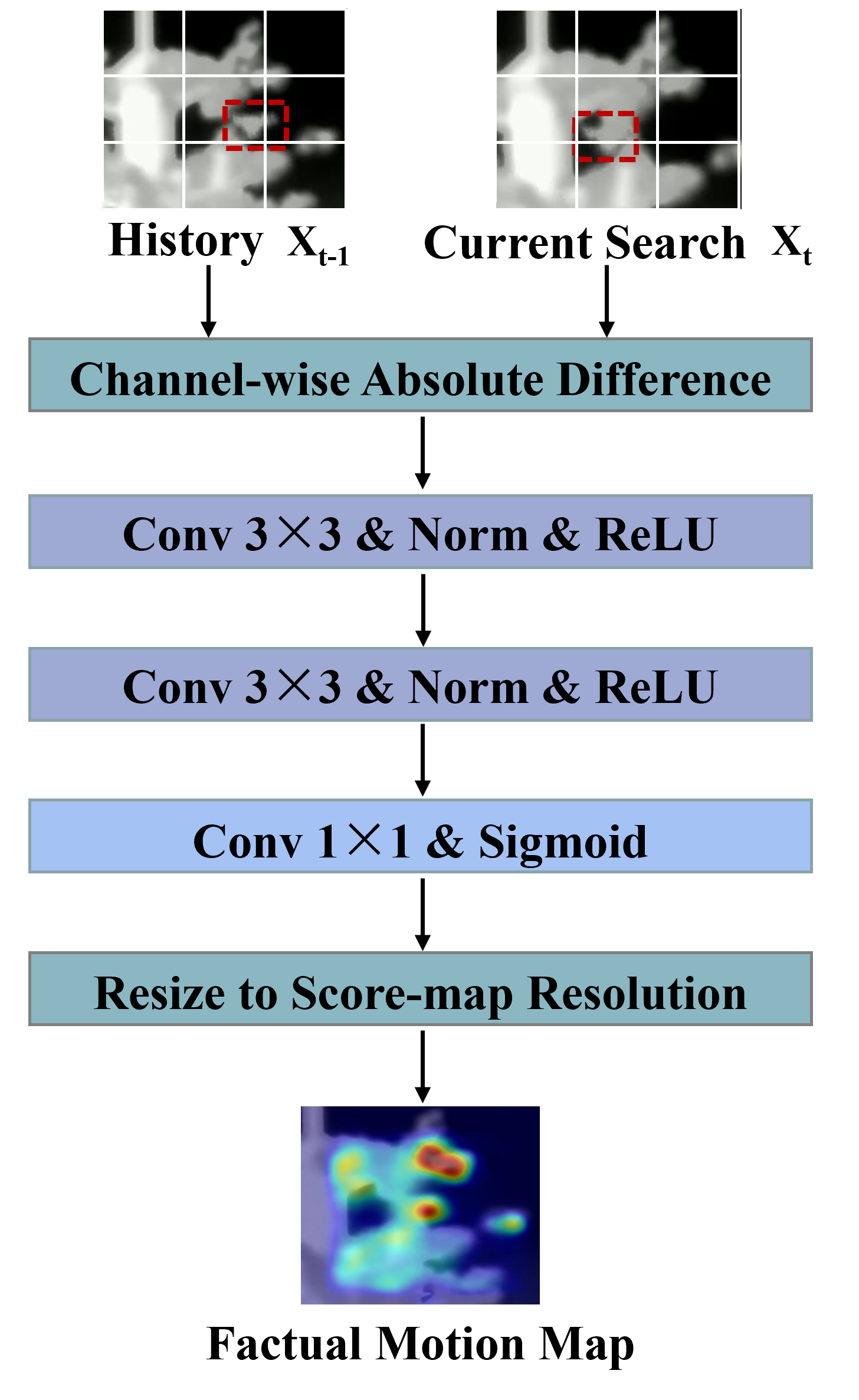}
    \caption{
    Illustration of the motion evidence encoder. Given the historical search region $X_{t-1}$ and the current search region $X_t$, the encoder first computes a channel-averaged absolute difference map. The difference map is then processed by two $3\times3$ convolutional blocks and a $1\times1$ prediction layer, and finally resized to the score-map resolution to generate the factual motion map $M_t$.
    }
    \label{fig:motion_encoder}
\end{figure}

\noindent\textbf{a. Factual Motion Evidence.}
As shown in Fig.~\ref{fig:motion_encoder}, we introduce a lightweight motion evidence encoder to extract temporal cues from adjacent search regions. Given the historical search region $X_{t-1}$ and the current search region $X_t$, we first compute a channel-averaged absolute difference map:
\begin{equation}
    D_t = \frac{1}{C}\sum_{c=1}^{C}\left|X_t^c - X_{t-1}^c\right|,
\end{equation}
where $C$ denotes the number of image channels. The difference map $D_t$ highlights pixel-level temporal variations between the current and historical observations.

The difference map is then fed into a lightweight convolutional motion encoder $\Phi$. Specifically, $\Phi$ consists of two $3\times3$ convolutional blocks, each followed by batch normalization and ReLU activation, and a final $1\times1$ convolution followed by a sigmoid function. The predicted response is resized to the score-map resolution, producing the factual motion map:
\begin{equation}
    M_t = \Phi(X_t, X_{t-1}),
\end{equation}
where $M_t \in [0,1]^{1\times H \times W}$ denotes the motion response map under the factual historical observation. This map provides explicit temporal evidence for the subsequent motion-guided token modulation and reliability-aware score fusion modules.

To provide direct supervision for the factual motion map, we generate a target heatmap $G_t$ from the ground-truth bounding box and apply a focal loss to $M_t$:
\begin{equation}
    \mathcal{L}_{motion} = \mathcal{L}_{focal}(M_t, G_t).
\end{equation}
This supervision encourages the motion encoder to produce high responses around the target region while suppressing irrelevant background responses.

\noindent\textbf{b. Counterfactual Target-Erased History.}
The factual motion map $M_t$ captures temporal variations between the current search region and the historical search region. However, strong temporal responses are not always caused by reliable target motion. In infrared UAV tracking, camera jitter, background dynamics, and thermal noise may also introduce apparent changes. To regularize the motion encoder with an explicit counterfactual reference, we construct a target-erased historical search region during training.

Given the historical search region $X_{t-1}$ and the target box $b_{t-1}$ in the historical frame, we erase a scaled target region from $X_{t-1}$:
\begin{equation}
    \tilde{X}_{t-1} = \mathcal{E}(X_{t-1}, \gamma b_{t-1}),
\end{equation}
where $\gamma$ is the erasing scale factor, and $\mathcal{E}(\cdot)$ replaces pixels inside the scaled box $\gamma b_{t-1}$ with the global mean value of the historical search region. The scaled erasing box slightly enlarges the target region to reduce residual target evidence around the boundary. The current search region $X_t$ is kept unchanged, since it is the observation to be localized.

We then apply the same motion encoder $\Phi$ to the current search region and the counterfactual history:
\begin{equation}
    \tilde{M}_t = \Phi(X_t, \tilde{X}_{t-1}),
\end{equation}
where $\tilde{M}_t$ denotes the counterfactual motion map. The target-erased history breaks the temporal consistency of the target trajectory while preserving the current observation and most background context. Therefore, $\tilde{M}_t$ serves as a hard counterfactual reference for motion reliability learning. This counterfactual branch is used only during training and is removed during inference.

\noindent\textbf{c. Counterfactual Motion Reliability Learning.}
The counterfactual motion map $\tilde{M}_t$ is not treated as a zero-response negative sample. Since the target may still appear in the current search region $X_t$, and background changes can also produce valid temporal differences, directly suppressing all counterfactual responses would introduce an overly strong and unrealistic constraint. Instead, we use $\tilde{M}_t$ as a hard counterfactual reference and learn motion reliability from the relative response gap between factual and counterfactual observations.

Let $G_t \in [0,1]^{1\times H \times W}$ denote the target foreground heatmap generated from the ground-truth box, and let $\bar{G}_t = 1-G_t$ denote the background region. We compute the average factual foreground response, counterfactual foreground response, factual background response, and counterfactual background response as
\begin{equation}
    s_f = \mathrm{Avg}(M_t \odot G_t), \quad
    \tilde{s}_f = \mathrm{Avg}(\tilde{M}_t \odot G_t),
\end{equation}
\begin{equation}
    s_b = \mathrm{Avg}(M_t \odot \bar{G}_t), \quad
    \tilde{s}_b = \mathrm{Avg}(\tilde{M}_t \odot \bar{G}_t),
\end{equation}
where $\odot$ denotes element-wise multiplication. Since $M_t$ is computed from the factual historical observation while $\tilde{M}_t$ is computed from the target-erased history, the factual foreground response is expected to be more reliable than the counterfactual foreground response. We therefore define a counterfactual foreground ranking loss:
\begin{equation}
    \mathcal{L}_{cf} = \max(0, m - s_f + \tilde{s}_f),
\end{equation}
where $m$ is the margin.

To further suppress background-induced pseudo motion, we require the factual foreground response to be larger than the strongest background response from both factual and counterfactual maps:
\begin{equation}
    \mathcal{L}_{cf-bg}
    =
    \max(0, m - s_f + \max(s_b, \tilde{s}_b)).
\end{equation}
These relative constraints encourage the motion encoder to assign high responses to target-consistent motion while reducing responses caused by background changes or noise.

In addition, we derive a soft reliability target from the gap between the factual foreground response and the strongest competing response:
\begin{equation}
    y_r =
    \sigma\left(
    \frac{s_f-\max(\tilde{s}_f, s_b)}{\tau}
    \right),
\end{equation}
where $\tau$ is a temperature parameter. The predicted reliability score $r_t$ from the score fusion module is supervised by
\begin{equation}
    \mathcal{L}_{rel} = \mathrm{BCE}(r_t, y_r).
\end{equation}
This reliability supervision links counterfactual motion learning with the final score fusion module, enabling the tracker to adaptively exploit motion cues when they are trustworthy.

Finally, the training objective for the motion evidence encoder is defined as
\begin{equation}
    \mathcal{L}_{MEE}
    =
    \mathcal{L}_{motion}
    +
    \mathcal{L}_{cf}
    +
    \mathcal{L}_{cf-bg}
    +
    \mathcal{L}_{rel}.
\end{equation}
All these losses are computed only on valid positive target samples to avoid invalid supervision from target-absent frames.

\subsection{Motion-Guided Token Modulation}

After obtaining the factual motion map $M_t$, we use it to enhance motion-relevant search tokens before the localization head. Let $E_X \in \mathbb{R}^{N \times C}$ denote the search tokens produced by the one-stream tracking backbone, where $N$ is the number of search tokens and $C$ is the feature dimension. The motion map $M_t$ is resized and flattened to match the spatial layout of the search tokens, resulting in a token-wise motion gate $\hat{M}_t \in \mathbb{R}^{N \times 1}$. We then apply a residual modulation operation:
\begin{equation}
    \tilde{E}_X = E_X \odot (1 + \alpha \hat{M}_t),
\end{equation}
where $\alpha$ controls the modulation strength and $\odot$ denotes element-wise multiplication.

The residual form preserves the original appearance representation while adaptively enhancing tokens located in motion-relevant regions. The guided search tokens $\tilde{E}_X$ are then fed into the localization head for target localization. In our implementation, only the search tokens are modulated, while the template tokens remain unchanged.

\subsection{Reliability-Aware Score Fusion}

In addition to modulating search tokens, the factual motion map $M_t$ is further used to refine the final target response. Given the appearance score map $S_t$ predicted by the box head and the motion map $M_t$ generated by the motion evidence encoder, we design a reliability-aware score fusion module to adaptively combine appearance and motion cues.

Specifically, we concatenate $S_t$ and $M_t$ along the channel dimension and feed them into a lightweight convolutional fusion module. This module first uses a $3\times3$ convolution followed by batch normalization and GELU activation to extract local appearance-motion interaction features. A subsequent $1\times1$ convolution predicts a single-channel residual correction. To stabilize the correction, the residual is passed through a hyperbolic tangent function and scaled by a residual factor. The resulting residual correction is denoted as $\Delta_t$:
\begin{equation}
    \Delta_t = \mathcal{F}_{res}([S_t, M_t]),
\end{equation}
where $\mathcal{F}_{res}(\cdot)$ denotes the residual fusion network and $[\cdot,\cdot]$ denotes channel-wise concatenation. Instead of directly replacing the original appearance response, we apply the residual correction in the logit space:
\begin{equation}
    S_t^{m} =
    \sigma\left(\mathrm{logit}(S_t) + \Delta_t\right),
\end{equation}
where $\mathrm{logit}(\cdot)$ maps the probability score to the logit space and $\sigma(\cdot)$ maps it back to the probability range.

However, motion evidence can be unreliable under camera jitter, background motion, and infrared noise. Therefore, we estimate a reliability score $r_t$ to control the contribution of motion-aware correction. We first compute the peak responses of the appearance score map and the motion map:
\begin{equation}
    p_s = \max(S_t), \quad p_m = \max(M_t).
\end{equation}
Then, we estimate their response centers using soft-argmax:
\begin{equation}
    c_s = \mathrm{SoftArgmax}(S_t), \quad
    c_m = \mathrm{SoftArgmax}(M_t).
\end{equation}
The normalized distance between the two centers is computed as
\begin{equation}
    d_{sm} = \frac{\|c_s - c_m\|_2}{\sqrt{2}}.
\end{equation}
The reliability score is predicted from these compact statistics:
\begin{equation}
    r_t =
    \mathrm{MLP}\left([p_s, p_m, d_{sm}]\right),
\end{equation}
where the MLP ends with a sigmoid activation, ensuring $r_t \in [0,1]$.

The final score map is computed as
\begin{equation}
    S_t^{final}
    =
    S_t + \beta r_t (S_t^{m} - S_t),
\end{equation}
where $\beta$ is a fusion coefficient. When the motion evidence is reliable, $r_t$ increases the contribution of the motion-aware score $S_t^{m}$. When motion evidence is uncertain, the correction is suppressed and $S_t^{final}$ remains close to the appearance score map $S_t$.

To optimize the fusion module, we supervise the final score map $S_t^{final}$ with the target heatmap $G_t$ using a focal loss:
\begin{equation}
    \mathcal{L}_{fusion}
    =
    \mathcal{L}_{focal}(S_t^{final}, G_t).
\end{equation}
This loss encourages the fused score map to preserve accurate target localization while benefiting from reliable motion cues. The reliability score $r_t$ is further supervised by the counterfactual reliability loss $\mathcal{L}_{rel}$ described in the motion reliability learning part. The final tracking output is represented by the predicted box $B_t$ and the motion-aware final score map $S_t^{final}$.

\subsection{Training and Inference}

After defining the objectives of the proposed modules, we summarize the overall training objective of CMRTrack. The basic tracking objective follows OSTrack and includes the standard localization losses for bounding box regression:
\begin{equation}
    \mathcal{L}_{track}
    =
    \lambda_{giou}\mathcal{L}_{giou}
    +
    \lambda_{1}\mathcal{L}_{1},
\end{equation}
where $\mathcal{L}_{giou}$ and $\mathcal{L}_{1}$ supervise the predicted bounding box $B_t$. Following the standard OSTrack setting, we set $\lambda_{giou}=2$ and $\lambda_{1}=5$.

The proposed motion evidence encoder is optimized by $\mathcal{L}_{MEE}$, which includes factual motion supervision, counterfactual motion constraints, and reliability supervision. The reliability-aware score fusion module is optimized by $\mathcal{L}_{fusion}$, which supervises the final score map $S_t^{final}$. The overall training objective is formulated as
\begin{equation}
    \mathcal{L}
    =
    \mathcal{L}_{track}
    +
    \mathcal{L}_{MEE}
    +
    \mathcal{L}_{fusion}.
\end{equation}
All motion-related losses are computed only on valid positive target samples to avoid invalid supervision from target-absent frames.

During inference, CMRTrack operates in a causal manner without using counterfactual inputs. Given the template $Z$, the current search region $X_t$, and the previous search region $X_{t-1}$, the motion encoder first produces the factual motion map $M_t=\Phi(X_t,X_{t-1})$. The one-stream backbone then extracts template-search features, where $M_t$ is used to modulate the search tokens before the localization head. The localization head predicts the bounding box $B_t$ and the appearance score map $S_t$, and the reliability-aware score fusion module further combines $S_t$ with $M_t$ to obtain the final score map $S_t^{final}$. After processing each frame, the current search region is stored as the historical observation for the next frame. Since the counterfactual target-erased branch is used only for training-time reliability learning, it is removed during inference and introduces no additional counterfactual forward pass.

\section{Experiments}

\subsection{Implementation Details}

CMRTrack is implemented in Python with PyTorch. Unless otherwise specified, all training and evaluation experiments are conducted on NVIDIA RTX 5090 GPUs. The tracker adopts a ViT-B backbone with a center-based localization head, following the one-stream tracking paradigm of OSTrack. The backbone is initialized with DropMAE pretrained weights, while the newly introduced motion-related modules, including the motion evidence encoder, motion-guided token modulation, and reliability-aware score fusion module, are randomly initialized.

The model is trained on the Anti-UAV410 training set. The template and search regions are cropped with search factors of 2.0 and 6.0, and resized to $128\times128$ and $256\times256$, respectively. During training, positive and negative samples are sampled with a ratio of 7:3. The model is trained for 30 epochs using the AdamW optimizer. The initial learning rate is set to $4\times10^{-4}$, and the backbone learning rate is multiplied by 0.1. The batch size is 64, and the weight decay is $10^{-4}$. The motion evidence encoder uses one historical search region and a hidden dimension of 16. The token modulation coefficient and score fusion coefficient are both set to 0.5.

For the counterfactual branch, the target-erased history is generated by replacing the scaled historical target region with the global mean value of the historical search region. During inference, CMRTrack follows a causal tracking pipeline, where the previous search region is maintained as the historical observation. The counterfactual branch is disabled during inference. Tracking performance is evaluated using standard Anti-UAV metrics, including AUC, precision, normalized precision, and state accuracy. For fair efficiency comparison with previous methods, the speed evaluation in the efficiency analysis is conducted on a single NVIDIA RTX 3090 GPU.

\subsection{State-of-the-Art Comparison}

\noindent\textbf{1) Evaluation on Anti-UAV.}
The Anti-UAV benchmark contains more than 318 video pairs with over 580K manually annotated bounding boxes, covering both visible and thermal infrared UAV tracking scenarios. In this work, we evaluate CMRTrack on the thermal infrared modality, which is more suitable for long-range UAV tracking under low illumination, weak target appearance, and complex background interference.

Table~\ref{tab:antiuav_results} reports the comparison between CMRTrack and representative state-of-the-art trackers on the Anti-UAV testing set. For a fair comparison, all reported methods are trained on the Anti-UAV410 training set. CMRTrack achieves the best overall performance, obtaining 72.3\% AUC, 92.5\% precision, 91.7\% normalized precision, and 73.6\% state accuracy. Compared with the strongest competing tracker, FocusTrack, CMRTrack improves AUC, normalized precision, and state accuracy by 4.6, 3.3, and 4.7 percentage points, respectively. These consistent improvements demonstrate that counterfactual motion reliability learning and reliability-aware score fusion effectively enhance localization accuracy and tracking robustness in thermal infrared UAV tracking.

\begin{table}[t]
\centering
\caption{Comparison with state-of-the-art methods on the Anti-UAV testing set. The best results are highlighted in bold.}
\label{tab:antiuav_results}
\resizebox{\linewidth}{!}{
\begin{tabular}{lcccccc}
\toprule
\multirow{2}{*}{Method} & \multirow{2}{*}{Source} & \multirow{2}{*}{Size} & \multicolumn{4}{c}{Anti-UAV} \\
\cmidrule(lr){4-7}
 &  &  & AUC & P & $P_{\mathrm{Norm}}$ & SA \\
\midrule
OSTrack~\cite{ye2022ostrack}       & ECCV 22  & 256 & 59.2 & 79.4 & 77.5 & 60.2 \\
ROMTrack~\cite{cai2023romtrack}    & ICCV 23  & 256 & 59.4 & 78.9 & 77.1 & 60.5 \\
ZoomTrack~\cite{kou2023zoomtrack}  & NeurIPS 23 & 256 & 63.5 & 86.0 & 83.4 & 64.5 \\
DropTrack~\cite{wu2023dropmae}   & CVPR 23  & 256 & 64.2 & 85.8 & 83.2 & 65.2 \\
FocusTrack~\cite{wang2025focustrack}        & TGRS 25     & 256 & 67.7 & 90.9 & 88.4 & 68.9 \\
\midrule
CMRTrack                            & Ours     & 256 & \textbf{72.3} & \textbf{92.5} & \textbf{91.7} & \textbf{73.6} \\
\bottomrule
\end{tabular}
}
\end{table}

\noindent\textbf{2) Evaluation on Anti-UAV410.}
We further compare CMRTrack with representative state-of-the-art trackers on the Anti-UAV410 testing set. Following the evaluation protocol in FocusTrack, the compared methods are divided into two groups. The upper section of Table~\ref{tab:sota_antiuav410} reports results without training on the Anti-UAV410 training set, while the lower section presents results after retraining on it. Tracking performance is evaluated using AUC, precision (P), normalized precision ($P_{norm}$), and state accuracy (SA).

As shown in Table~\ref{tab:sota_antiuav410}, CMRTrack achieves the best performance among all compared trackers. Compared with the OSTrack baseline, CMRTrack improves AUC from 53.7\% to 67.3\%, precision from 73.9\% to 89.9\%, normalized precision from 70.9\% to 86.2\%, and SA from 54.7\% to 68.5\%. The absolute gains are 13.6, 16.0, 15.3, and 13.8 percentage points, respectively. Compared with the strong Anti-UAV tracker FocusTrack, CMRTrack also achieves clear improvements of 4.5, 3.7, 3.4, and 4.6 percentage points in AUC, precision, normalized precision, and SA, respectively. These results indicate that reliable temporal motion evidence provides effective complementary information to appearance-based tracking.

\begin{figure*}[t]
    \centering
    \includegraphics[width=\linewidth]{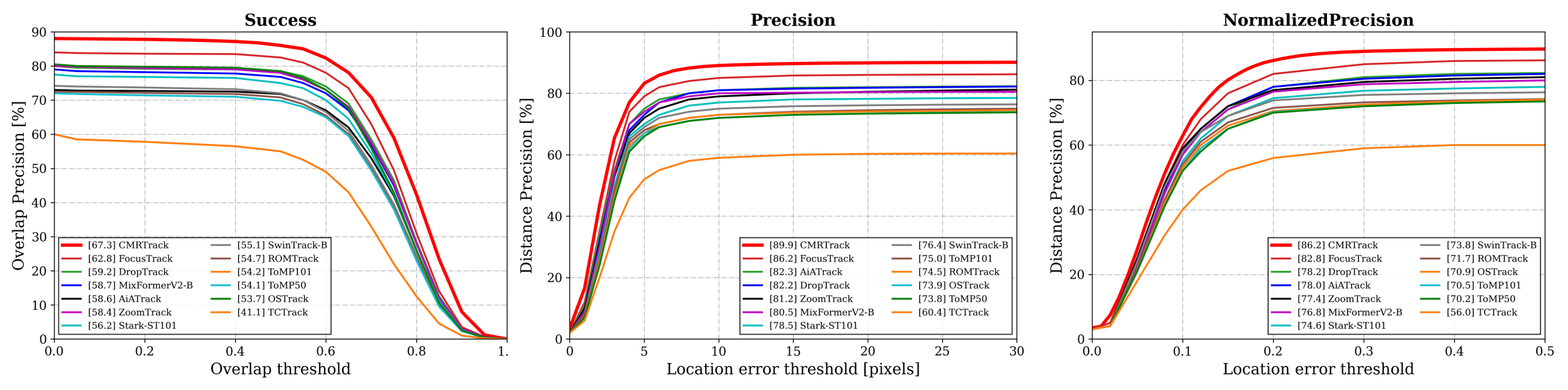}
    \caption{
    Success, precision, and normalized precision plots on the Anti-UAV410 testing set.
    CMRTrack achieves the best performance on all three evaluation curves, demonstrating its effectiveness in both overlap-based localization and center-based precision evaluation.
    }
    \label{fig:spn_antiuav410}
\end{figure*}

Fig.~\ref{fig:spn_antiuav410} shows the success, precision, and normalized precision plots on the Anti-UAV410 testing set. CMRTrack consistently achieves the best performance across all three curves. In the success plot, our method maintains a clear advantage under strict overlap thresholds, indicating more accurate bounding box localization. In the precision and normalized precision plots, CMRTrack also remains above competing methods, demonstrating improved center-location accuracy and scale-normalized robustness. Overall, the quantitative results verify the effectiveness of counterfactual motion reliability learning for challenging infrared UAV tracking.

\begin{table}[t]
    \centering
    \caption{
    Comparison with state-of-the-art trackers on the Anti-UAV410 testing set.
    The upper section reports results without training on the Anti-UAV410 training set, while the lower section presents results after retraining on it.
    The best and second-best results are marked in \textcolor{red}{red} and \textcolor{blue}{blue}, respectively.
    }
    \label{tab:sota_antiuav410}
    \setlength{\tabcolsep}{4pt}
    \begin{tabular}{lcccccc}
        \toprule
        \multirow{2}{*}{Method} & \multirow{2}{*}{Source} & \multirow{2}{*}{Size}
        & \multicolumn{4}{c}{Anti-UAV410} \\
        \cmidrule(lr){4-7}
        & & & AUC & P & $P_{norm}$ & SA \\
        \midrule
        TransT & CVPR 21 & 256 & 48.2 & 67.7 & 64.1 & 48.9 \\
        ETTrack & WACV 23 & 256 & 41.5 & 59.7 & 54.8 & 41.6 \\
        MixFormerV2-S & NeurIPS 23 & 224 & 45.6 & 64.1 & 60.0 & 46.1 \\
        GRM & CVPR 23 & 256 & 42.3 & 58.5 & 55.1 & 42.2 \\
        ARTrack & CVPR 23 & 256 & 48.2 & 67.2 & 62.9 & 48.5 \\
        JointNLT & CVPR 23 & 320 & 48.4 & 69.0 & 64.5 & 48.9 \\
        SeqTrack & CVPR 23 & 256 & 52.2 & 73.8 & 70.0 & 52.9 \\
        PromptVT & TCSVT 24 & 320 & 50.5 & 71.5 & 65.6 & 51.2 \\
        \midrule
        Stark-ST101 & ICCV 21 & 320 & 56.2 & 78.5 & 74.6 & 57.1 \\
        TCTrack & CVPR 22 & 287 & 41.1 & 60.4 & 56.0 & 41.6 \\
        ToMP50 & CVPR 22 & 288 & 54.1 & 73.8 & 70.2 & 55.1 \\
        ToMP101 & CVPR 22 & 288 & 54.2 & 75.0 & 70.5 & 55.1 \\
        SwinTrack-Tiny & NeurIPS 22 & 224 & 53.0 & 71.4 & 68.1 & 53.1 \\
        SwinTrack-Base & NeurIPS 22 & 384 & 55.9 & 76.4 & 72.3 & 55.7 \\
        AiATrack & ECCV 22 & 320 & 58.6 & 82.3 & 78.0 & 59.6 \\
        ROMTrack & ICCV 23 & 256 & 54.7 & 74.5 & 71.7 & 55.7 \\
        ZoomTrack & NeurIPS 23 & 256 & 58.4 & 81.2 & 77.4 & 59.4 \\
        MixFormerV2-B & NeurIPS 23 & 288 & 58.7 & 80.5 & 76.8 & 59.6 \\
        DropTrack & CVPR 23 & 256 & 59.2 & 82.2 & 78.2 & 60.2 \\
        FocusTrack & TGRS 25 & 256
        & \textcolor{blue}{62.8}
        & \textcolor{blue}{86.2}
        & \textcolor{blue}{82.8}
        & \textcolor{blue}{63.9} \\
        \midrule
        OSTrack & Baseline & 256 & 53.7 & 73.9 & 70.9 & 54.7 \\
        CMRTrack & Ours & 256
        & \textcolor{red}{67.3}
        & \textcolor{red}{89.9}
        & \textcolor{red}{86.2}
        & \textcolor{red}{68.5} \\
        \bottomrule
    \end{tabular}
\end{table}

\subsection{Attribute-Based Analysis}

To comprehensively evaluate the robustness of CMRTrack, we conduct attribute-based analysis on the Anti-UAV410 testing set. Six challenging attributes are considered, including dynamic background clutter (DBC), scale variation (SV), fast motion (FM), occlusion (OC), thermal crossover (TC), and out-of-view (OV). We also evaluate trackers under different target-scale subsets, including normal-size, medium-size, small-size, and tiny-size targets. Radar plots, success plots, and precision plots are reported for a comprehensive comparison.

\begin{figure}[t]
    \centering
    \includegraphics[width=\linewidth]{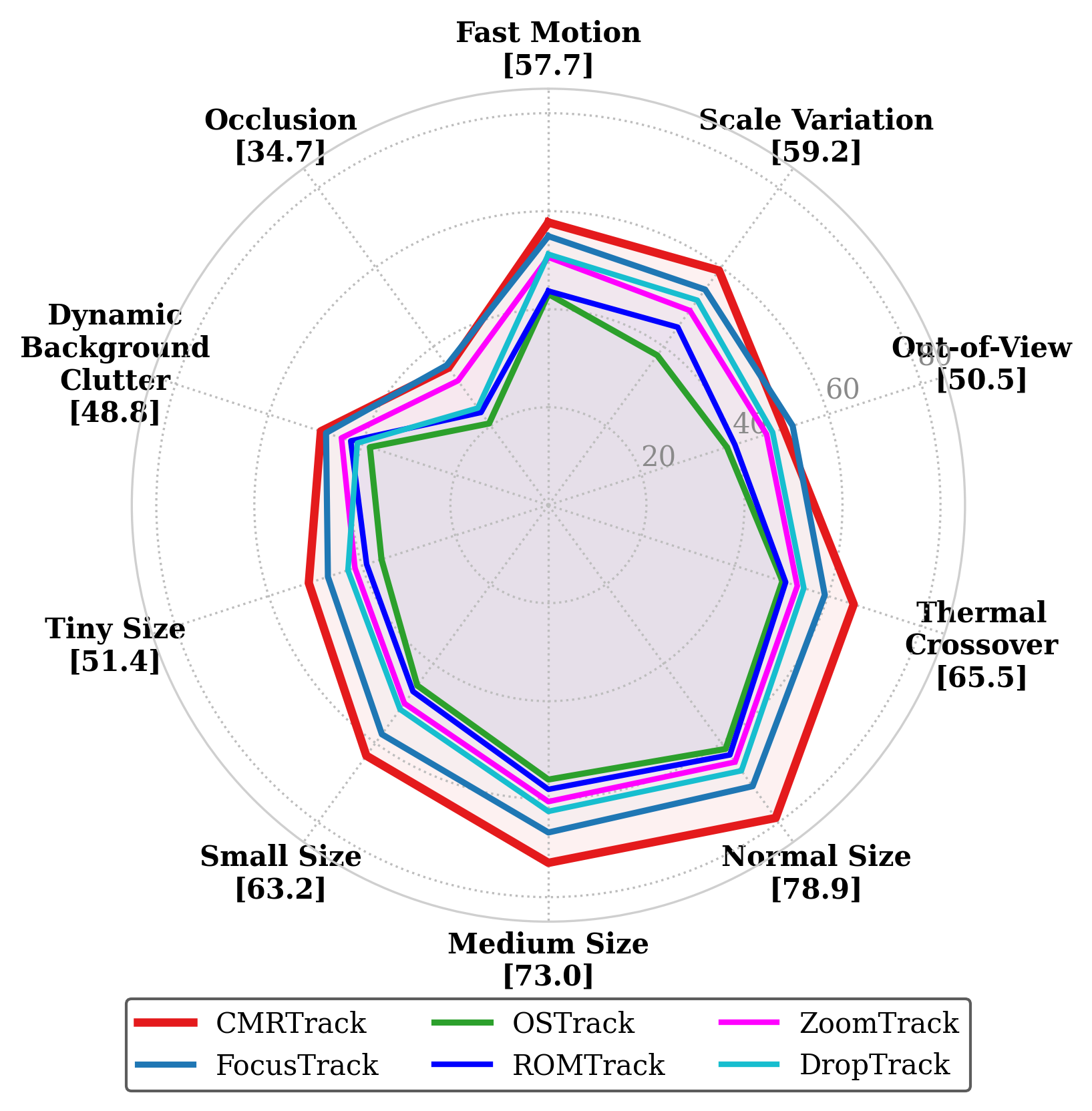}
    \caption{
    Radar plot of attribute-based success performance on the Anti-UAV410 testing set. The radar chart summarizes challenging attributes and target-scale attributes, including FM, SV, OV, TC, normal size, medium size, small size, tiny size, DBC, and OC.
    }
    \label{fig:attribute_radar}
\end{figure}

Fig.~\ref{fig:attribute_radar} provides an overall comparison across challenging attributes and target scales. CMRTrack forms the outermost polygon on most axes, indicating balanced robustness under diverse infrared UAV scenarios. In particular, CMRTrack achieves 57.7\%, 59.2\%, 65.5\%, and 48.8\% AUC on FM, SV, TC, and DBC, respectively, outperforming the OSTrack baseline by 14.6, 21.4, 15.3, and 10.5 percentage points. For target-scale subsets, CMRTrack obtains 78.9\%, 73.0\%, 63.2\%, and 51.4\% AUC on normal-size, medium-size, small-size, and tiny-size targets, improving over FocusTrack by 8.0, 6.2, 5.4, and 4.1 percentage points, respectively. These results show that the proposed motion reliability modeling is effective for both challenging motion/background conditions and small-scale UAV targets.

\begin{figure*}[t]
    \centering
    \includegraphics[width=\linewidth]{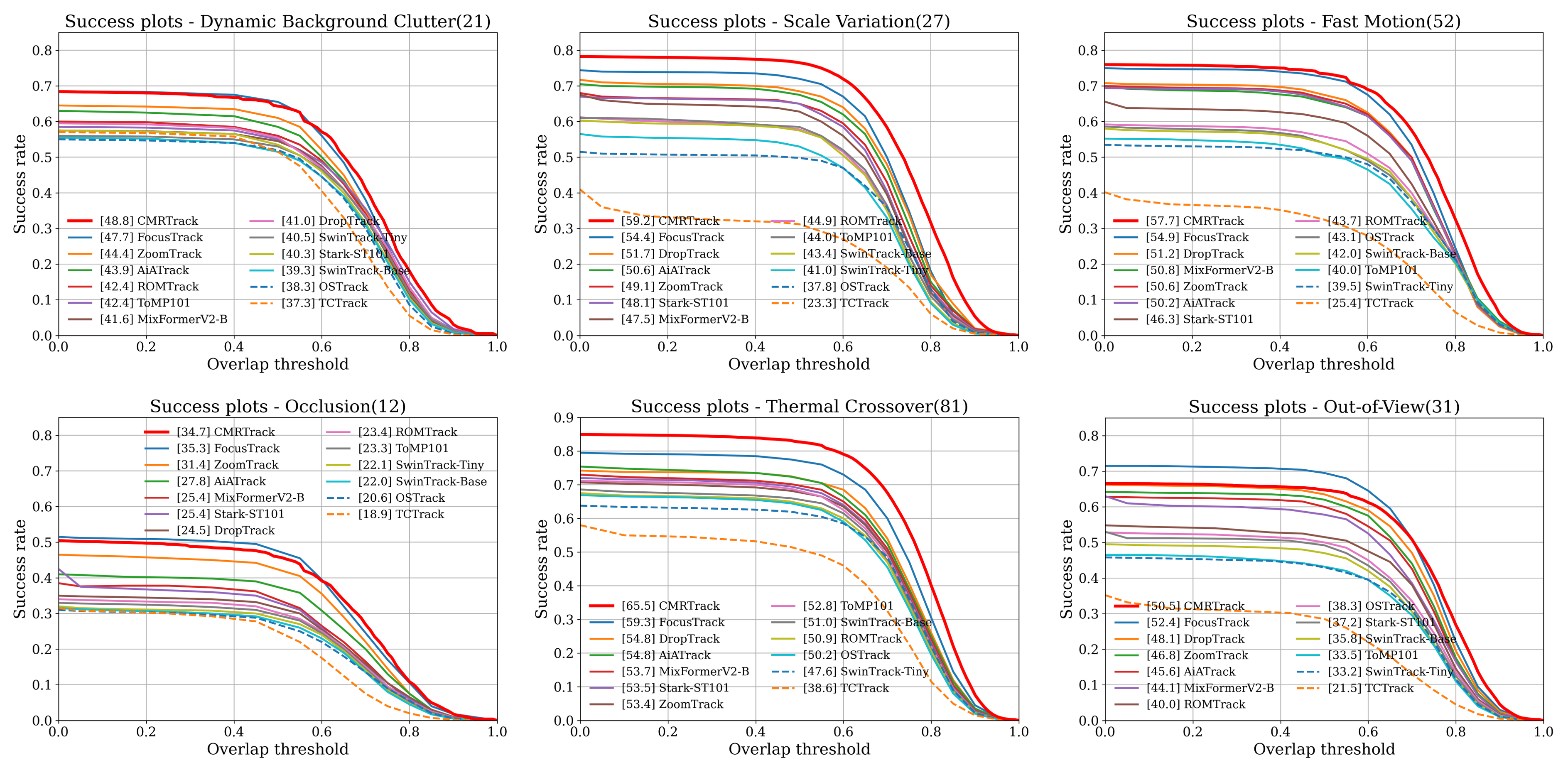}
    \caption{
    Attribute-based success plots on the Anti-UAV410 testing set, including dynamic background clutter (DBC), scale variation (SV), fast motion (FM), occlusion (OC), thermal crossover (TC), and out-of-view (OV).
    }
    \label{fig:attribute_success}
\end{figure*}

\begin{figure*}[htbp]
    \centering
    \includegraphics[width=\linewidth]{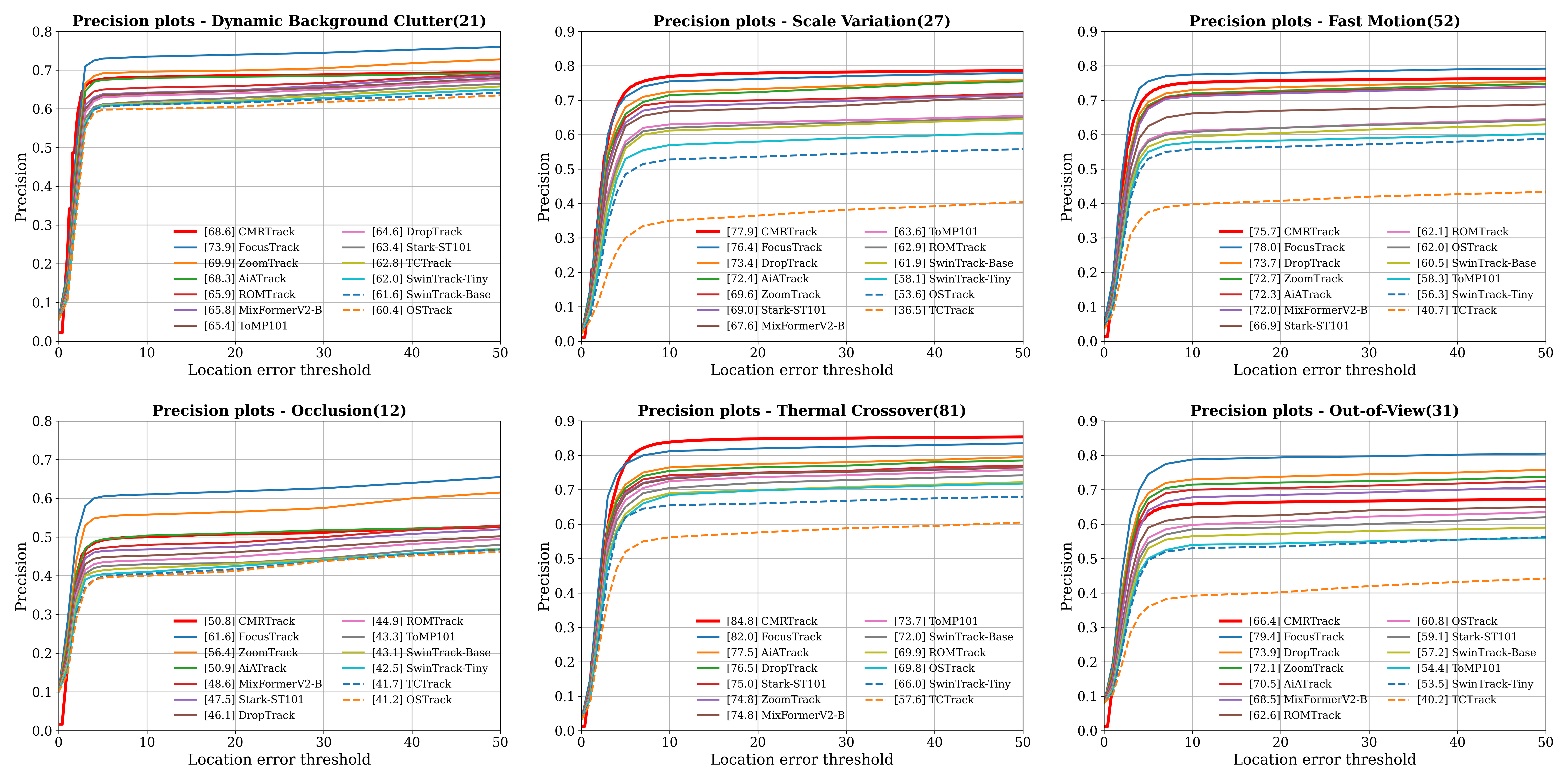}
    \caption{
    Attribute-based precision plots on the Anti-UAV410 testing set, including dynamic background clutter (DBC), scale variation (SV), fast motion (FM), occlusion (OC), thermal crossover (TC), and out-of-view (OV).
    }
    \label{fig:attribute_precision}
\end{figure*}

Fig.~\ref{fig:attribute_success} presents the success plots under six challenging attributes. CMRTrack achieves the best success performance on DBC, SV, FM, and TC, demonstrating strong overlap-based localization robustness against background interference, scale variation, rapid motion, and thermal distractors. The gains on SV and FM indicate that temporal motion evidence helps capture displacement cues when the target location changes rapidly. The improvements on DBC and TC further show that counterfactual motion reliability learning and reliability-aware score fusion can suppress unreliable responses caused by background changes and infrared distractors.

Fig.~\ref{fig:attribute_precision} reports the corresponding precision plots. CMRTrack achieves the best precision on SV and TC, and remains highly competitive on DBC and FM, indicating improved center-location accuracy. Although FocusTrack performs better in several disappearance-related scenarios such as OC and OV, CMRTrack consistently outperforms the OSTrack baseline across all attributes. Overall, the attribute-based results demonstrate that CMRTrack effectively enhances the robustness of one-stream tracking under fast motion, scale variation, thermal ambiguity, and dynamic infrared backgrounds.

\subsection{Counterfactual Erasing Strategy Analysis}

To analyze the influence of different target-erasing strategies, we compare several ways of constructing the counterfactual history on the Anti-UAV410 testing set. The results are reported in Table~\ref{tab:erasing_strategy}. All variants use the same training and evaluation settings, and only the target-erasing strategy is changed.

\begin{table}[htbp]
    \centering
    \caption{
    Analysis of different counterfactual target-erasing strategies on the Anti-UAV410 testing set.
    }
    \label{tab:erasing_strategy}
    \setlength{\tabcolsep}{0pt}
    \renewcommand{\arraystretch}{1.05}
    \begin{tabular*}{\columnwidth}{@{\extracolsep{\fill}}lcccc@{}}
        \toprule
        Erasing Strategy & AUC & P & $P_{\mathrm{norm}}$ & SA \\
        \midrule
        Zero Fill & 65.8 & 88.0 & 84.6 & 66.9 \\
        Random Noise Fill & 65.4 & 87.5 & 84.0 & 66.5 \\
        Gaussian Blur & 66.2 & 88.5 & 85.1 & 67.4 \\
        Local Mean Fill & 66.7 & 89.1 & 85.6 & 67.9 \\
        Global Mean Fill & \textbf{67.3} & \textbf{89.9} & \textbf{86.2} & \textbf{68.5} \\
        \bottomrule
    \end{tabular*}
\end{table}

As shown in Table~\ref{tab:erasing_strategy}, different erasing strategies have a clear influence on motion reliability learning. Directly filling the target region with zeros or random noise leads to inferior performance. Zero filling introduces unnatural intensity discontinuities, while random noise may generate noisy temporal differences that are mistakenly treated as motion evidence. Gaussian blur reduces abrupt artifacts but may still preserve the thermal structure of the target, making the counterfactual reference less effective.

Local mean filling achieves better performance by producing a more natural erased region. However, it can still be affected by local thermal distractors or residual target responses around the object. In contrast, global mean filling achieves the best performance across all metrics. It removes target-specific evidence while avoiding strong artificial motion artifacts, providing a stable and simple counterfactual reference for learning reliable target-consistent motion.

\subsection{Ablation Study}

Table~\ref{tab:ablation_components} reports the ablation results of the proposed components on the Anti-UAV410 testing set. The complete CMRTrack achieves the best performance with 67.3\% AUC, 89.9\% precision, 86.2\% normalized precision, and 68.5\% SA. Compared with the OSTrack baseline, CMRTrack improves AUC by 13.6 percentage points, demonstrating the overall effectiveness of introducing reliable motion modeling into the one-stream tracking framework.

The contribution of counterfactual motion reliability learning is verified by removing the counterfactual branch. Without counterfactual learning, AUC drops from 67.3\% to 63.0\%, indicating that the target-erased history provides useful reliability supervision for learning target-consistent motion. The effectiveness of motion utilization is further validated by removing motion-guided token modulation and reliability-aware score fusion. Without token modulation, AUC decreases to 62.8\%, showing that injecting motion evidence into search tokens is important for target-aware representation learning. Without score fusion, AUC decreases to 62.7\%, indicating that motion-aware response refinement contributes to final localization robustness.

The largest degradation among the proposed variants is observed when reliability supervision is removed, where AUC drops to 60.3\%. This confirms that explicitly learning motion reliability is crucial for suppressing unreliable motion responses caused by background dynamics and infrared noise. Overall, the ablation results verify the necessity of each proposed component and demonstrate that counterfactual motion reliability learning provides an important additional gain by improving the reliability of motion cues.

\begin{table}[htbp]
    \centering
    \caption{
    Ablation study of different components in CMRTrack on the Anti-UAV410 testing set.
    CMRTrack and its ablated variants are reported with the final checkpoint.
    }
    \label{tab:ablation_components}
    \setlength{\tabcolsep}{4pt}
    \begin{tabular}{lccccc}
        \toprule
        Method & AUC & $\Delta$AUC & P & $P_{norm}$ & SA \\
        \midrule
        OSTrack Baseline & 53.7 & -13.6 & 73.9 & 70.9 & 54.7 \\
        w/o Counterfactual Learning & 63.0 & -4.3 & 85.9 & 82.0 & 64.1 \\
        w/o Token Modulation & 62.8 & -4.5 & 85.6 & 81.7 & 63.9 \\
        w/o Score Fusion & 62.7 & -4.6 & 85.9 & 81.9 & 63.7 \\
        w/o Reliability Supervision & 60.3 & -7.0 & 81.7 & 78.4 & 61.4 \\
        \midrule
        CMRTrack & \textbf{67.3} & -- & \textbf{89.9} & \textbf{86.2} & \textbf{68.5} \\
        \bottomrule
    \end{tabular}
\end{table}

\subsection{Motion Baseline Comparison}

To further verify that the performance gain does not simply come from adding temporal differences, we compare CMRTrack with several motion-based baseline variants. The results are shown in Table~\ref{tab:motion_baseline}. All variants are built upon the same OSTrack baseline and trained under the same setting.

\begin{table}[htbp]
    \centering
    \caption{
    Comparison with different motion-based baseline variants on the Anti-UAV410 testing set.
    }
    \label{tab:motion_baseline}
    \setlength{\tabcolsep}{0pt}
    \renewcommand{\arraystretch}{1.05}
    \begin{tabular*}{\columnwidth}{@{\extracolsep{\fill}}lcccc@{}}
        \toprule
        Variant & AUC & P & $P_{\mathrm{norm}}$ & SA \\
        \midrule
        OSTrack & 53.7 & 73.9 & 70.9 & 54.7 \\
        + Raw Temporal Difference & 56.4 & 77.2 & 74.0 & 57.3 \\
        + Supervised Motion Map & 60.8 & 82.9 & 79.5 & 61.9 \\
        + Motion Fusion w/o Reliability & 62.7 & 85.9 & 81.9 & 63.7 \\
        CMRTrack & \textbf{67.3} & \textbf{89.9} & \textbf{86.2} & \textbf{68.5} \\
        \bottomrule
    \end{tabular*}
\end{table}

As shown in Table~\ref{tab:motion_baseline}, directly introducing raw temporal differences brings only limited improvement over the OSTrack baseline. This indicates that naive frame-level temporal changes are insufficient for robust infrared UAV tracking, since they may contain strong background variations and sensor noise. Adding a supervised motion map further improves performance, showing that explicit motion supervision is helpful for extracting target-related temporal cues. However, the variant with motion fusion but without reliability modeling still lags behind CMRTrack, suggesting that motion responses cannot be blindly trusted in complex infrared scenes.

Compared with these motion-based baselines, CMRTrack achieves the best performance across all metrics. The improvement demonstrates that the key factor is not merely introducing temporal differences or motion fusion, but explicitly learning motion reliability through counterfactual target-erased history. By distinguishing target-consistent motion from background-induced pseudo motion, CMRTrack can exploit useful temporal cues while suppressing unreliable motion responses.

\subsection{Parameter Analysis}

To evaluate the sensitivity of CMRTrack to key hyperparameters, we conduct parameter analysis on the Anti-UAV410 testing set. We vary one parameter at a time while keeping the others fixed to the default setting. The default setting is $\gamma=1.0$, $\alpha=0.5$, $\beta=0.5$ and $m=0.12$.

\begin{table}[htbp]
    \centering
    \caption{
    Parameter analysis of CMRTrack on the Anti-UAV410 testing set.
    The default setting is marked in bold.
    }
    \label{tab:param_analysis}
    \setlength{\tabcolsep}{0pt}
    \renewcommand{\arraystretch}{1}
    \begin{tabular*}{\columnwidth}{@{\extracolsep{\fill}}lccccc@{}}
        \toprule
        Parameter & Value & AUC & P & $P_{\mathrm{norm}}$ & SA \\
        \midrule
        $\gamma$ & 0.8 & 66.4 & 88.8 & 85.1 & 67.5 \\
        $\gamma$ & \textbf{1.0} & \textbf{67.3} & \textbf{89.9} & \textbf{86.2} & \textbf{68.5} \\
        $\gamma$ & 1.2 & 66.9 & 89.3 & 85.7 & 68.1 \\
        $\gamma$ & 1.5 & 66.1 & 88.5 & 84.9 & 67.2 \\
        \midrule
        $\alpha$ & 0.3 & 66.6 & 89.0 & 85.4 & 67.7 \\
        $\alpha$ & \textbf{0.5} & \textbf{67.3} & \textbf{89.9} & \textbf{86.2} & \textbf{68.5} \\
        $\alpha$ & 0.7 & 66.8 & 89.2 & 85.6 & 68.0 \\
        $\alpha$ & 1.0 & 65.9 & 88.1 & 84.6 & 66.9 \\
        \midrule
        $\beta$ & 0.3 & 66.7 & 89.1 & 85.5 & 67.8 \\
        $\beta$ & \textbf{0.5} & \textbf{67.3} & \textbf{89.9} & \textbf{86.2} & \textbf{68.5} \\
        $\beta$ & 0.7 & 66.9 & 89.4 & 85.8 & 68.1 \\
        $\beta$ & 1.0 & 66.0 & 88.4 & 84.8 & 67.1 \\
        \midrule
        $m$ & 0.08 & 66.6 & 89.0 & 85.3 & 67.6 \\
        $m$ & \textbf{0.12} & \textbf{67.3} & \textbf{89.9} & \textbf{86.2} & \textbf{68.5} \\
        $m$ & 0.16 & 66.8 & 89.3 & 85.6 & 68.0 \\
        $m$ & 0.20 & 66.2 & 88.7 & 85.0 & 67.3 \\
        \bottomrule
    \end{tabular*}
\end{table}

As shown in Table~\ref{tab:param_analysis}, CMRTrack is relatively stable within a reasonable range of hyperparameters. For the erasing scale $\gamma$, the best performance is obtained when $\gamma=1.0$. A smaller erasing region may leave residual target evidence in the historical search region, while a larger erasing region may remove excessive background context, leading to degraded reliability learning. For the token modulation coefficient $\alpha$, moderate modulation achieves the best result. When $\alpha$ is too small, motion evidence is insufficiently injected into search tokens; when it is too large, motion responses may over-amplify uncertain regions and disturb the original appearance representation.

A similar trend can be observed for the score fusion coefficient $\beta$. The default value $\beta=0.5$ achieves the best balance between appearance response and motion-aware correction. Excessive fusion strength may introduce unreliable motion responses into the final score map, while a smaller value limits the contribution of motion evidence. For the counterfactual margin $m$, CMRTrack performs best at $m=0.12$, indicating that a moderate ranking constraint is effective for separating factual target motion from counterfactual and background responses. Overall, the parameter analysis demonstrates that the proposed method is not overly sensitive to hyperparameter choices, and the default setting provides a favorable balance between motion enhancement and reliability control.

\subsection{Qualitative Analysis}

\begin{figure*}[t]
    \centering
    \includegraphics[width=\linewidth]{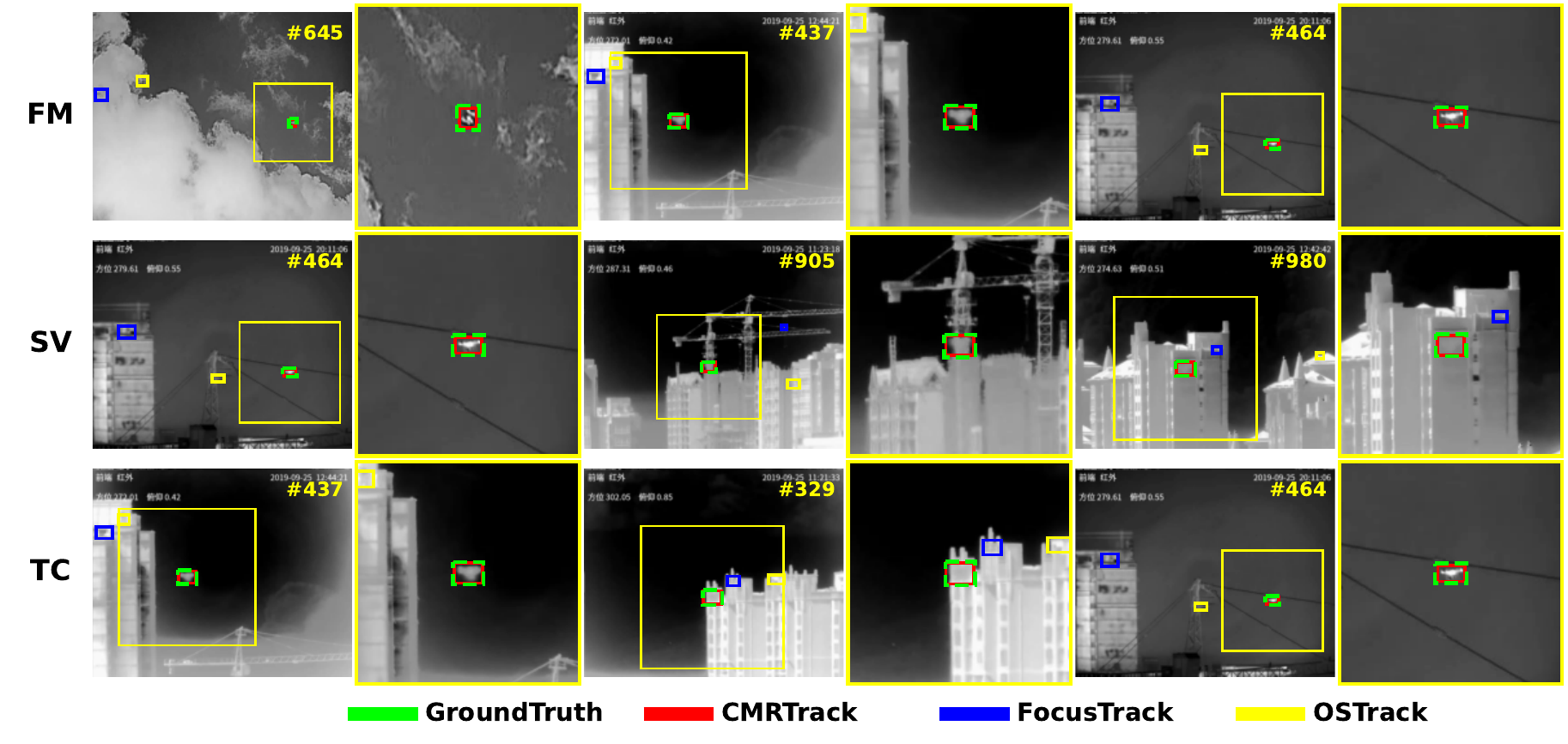}
    \caption{
    Qualitative comparison under FM, SV, and TC scenarios. The zoomed-in views show the local target regions.
    }
    \label{fig:qualitative_fm_sv_tc}
\end{figure*}

\begin{figure*}[htbp]
    \centering
    \includegraphics[width=\linewidth]{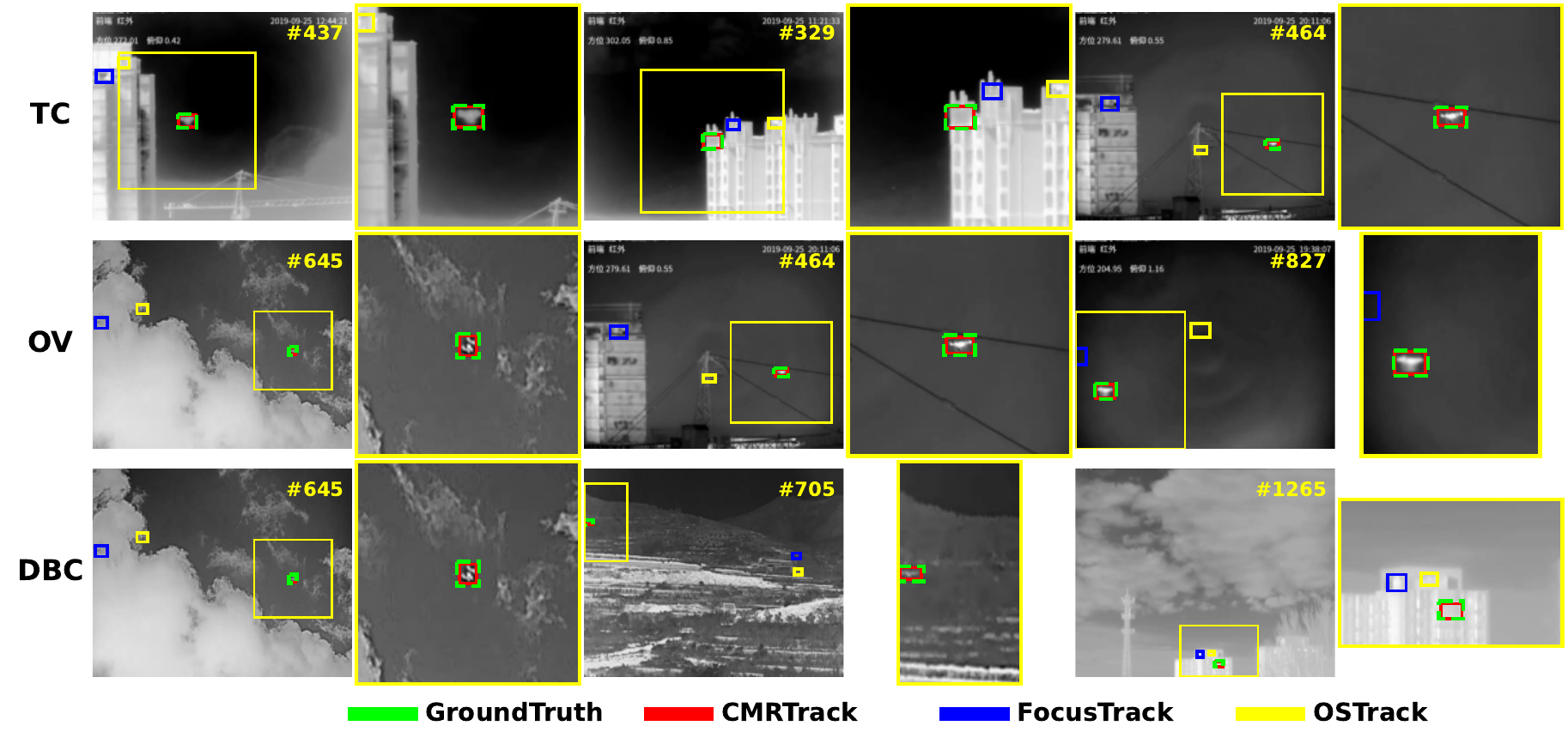}
    \caption{
    Qualitative comparison under TC, OV, and DBC scenarios. The zoomed-in views show the local target regions.
    }
    \label{fig:qualitative_tc_ov_dbc}
\end{figure*}

\noindent\textbf{1) Tracking Result Visualization.}
We visualize representative tracking results of CMRTrack, FocusTrack, and OSTrack under challenging infrared UAV scenarios, including fast motion (FM), scale variation (SV), thermal crossover (TC), out-of-view (OV), and dynamic background clutter (DBC). As shown in Figs.~\ref{fig:qualitative_fm_sv_tc} and~\ref{fig:qualitative_tc_ov_dbc}, the zoomed-in views highlight the local target regions for detailed comparison.

\begin{figure*}[htbp]
    \centering
    \includegraphics[width=\linewidth]{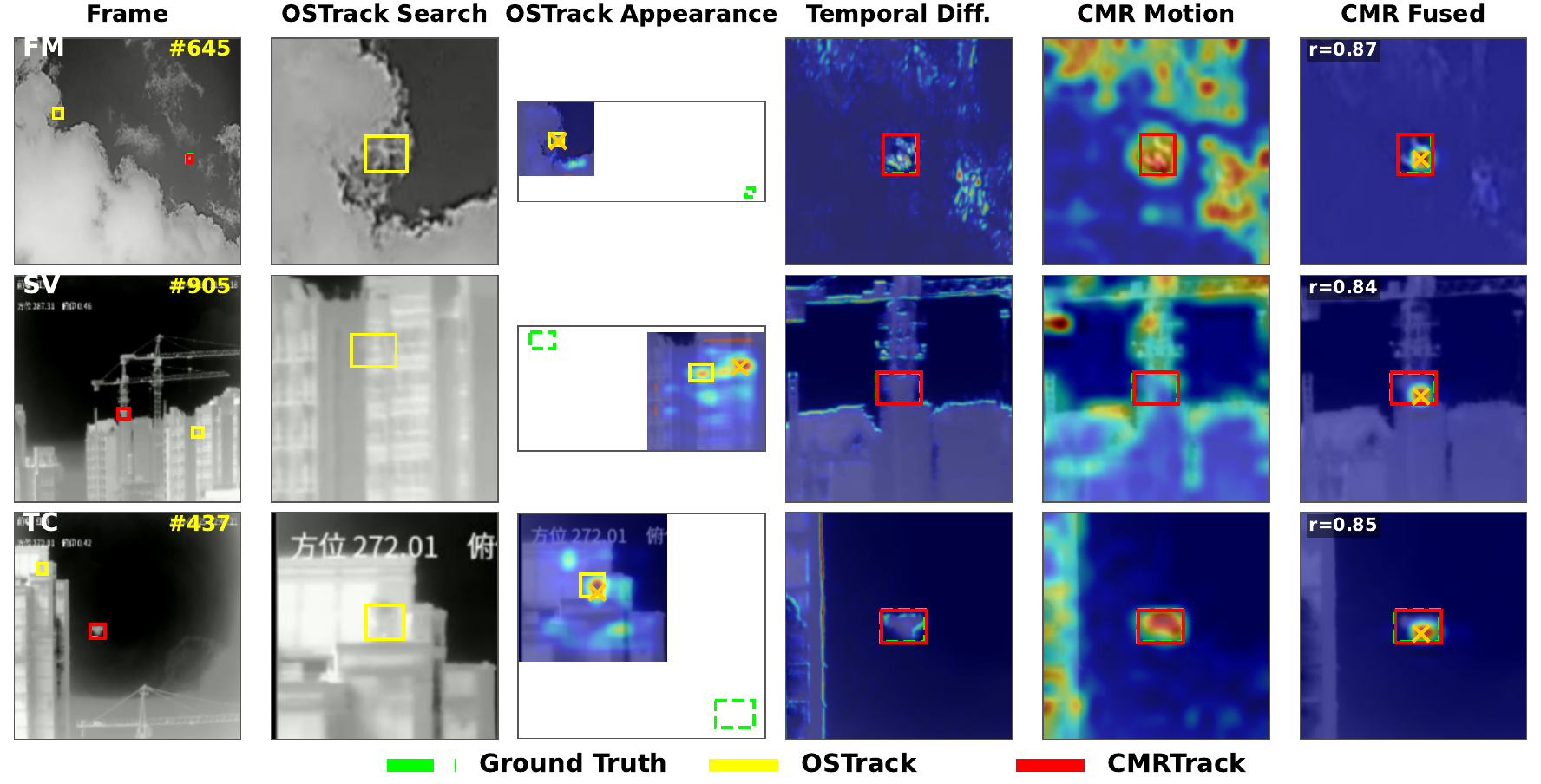}
    \caption{
    Intermediate response visualization of OSTrack and CMRTrack. OSTrack appearance responses can be attracted by background clutter or thermal distractors, while CMRTrack exploits temporal difference, motion evidence, and reliability-aware fusion to produce more target-consistent responses.
    }
    \label{fig:intermediate_response}
\end{figure*}

In FM and SV scenarios, abrupt displacement and scale changes make appearance-based trackers prone to drift. In TC and DBC scenarios, thermal distractors and background structures may produce target-like responses. Compared with OSTrack and FocusTrack, CMRTrack maintains more accurate localization across these cases, indicating that the proposed counterfactual motion reliability learning can enhance target-consistent temporal evidence while suppressing unreliable background responses. In OV cases, CMRTrack also shows better recovery when the target becomes partially invisible or reappears after severe displacement.

\noindent\textbf{2) Intermediate Response Visualization.}
To further explain the motion-aware design, we visualize intermediate responses of OSTrack and CMRTrack in Fig.~\ref{fig:intermediate_response}. We compare the OSTrack appearance response with the temporal difference, learned motion response, and final fused response of CMRTrack. Yellow boxes and crosses denote OSTrack predictions and response peaks, while red ones denote CMRTrack results.

As shown in Fig.~\ref{fig:intermediate_response}, OSTrack may generate response peaks on background structures with similar infrared intensity, especially under scale variation and thermal crossover. In contrast, CMRTrack converts adjacent-frame temporal differences into learned motion evidence and produces a fused response concentrated around the ground-truth target. This indicates that the proposed reliability-aware fusion selectively exploits target-consistent motion cues while suppressing unreliable appearance or background-induced motion responses.

\begin{figure}[t]
    \centering
    \includegraphics[width=\linewidth]{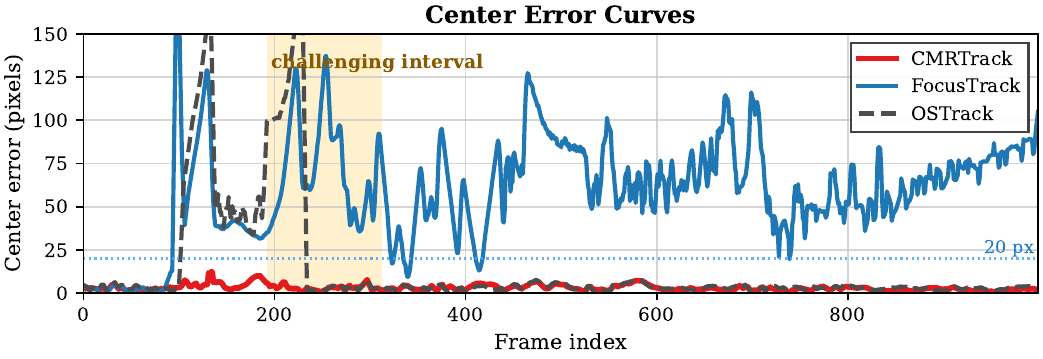}
    \caption{
    Center error curves on a representative challenging sequence. CMRTrack maintains consistently lower localization errors than FocusTrack and OSTrack.
    }
    \label{fig:center_error_curve}
\end{figure}

\noindent\textbf{3) Temporal Stability Analysis.}
We further evaluate temporal stability by plotting center error curves on a representative challenging sequence. As shown in Fig.~\ref{fig:center_error_curve}, the highlighted interval contains severe appearance and motion changes.

During this interval, FocusTrack and OSTrack show large center-error peaks, indicating temporary drift or inaccurate localization. In contrast, CMRTrack keeps lower errors and quickly recovers after local disturbances. These results show that the proposed counterfactual motion reliability learning and reliability-aware score fusion improve both frame-level localization and temporal stability under challenging infrared UAV scenarios.

\subsection{Efficiency Comparison}

Table~\ref{tab:efficiency_comparison} compares CMRTrack with representative local and global trackers in terms of tracking speed, computational complexity, and tracking performance. All speed results are measured on a single NVIDIA RTX 3090 GPU using the actual tracking process. MACs are computed with a $128\times128$ template and a $256\times256$ search region.

\begin{table}[htbp]
    \centering
    \caption{
    Efficiency comparison with representative trackers. Speed is measured on a single NVIDIA RTX 3090 GPU using the actual tracking process. MACs are computed with a $128\times128$ template and a $256\times256$ search region. The best results are highlighted in bold.
    }
    \label{tab:efficiency_comparison}
    \setlength{\tabcolsep}{3pt}
    \renewcommand{\arraystretch}{1.05}
    \resizebox{\columnwidth}{!}{%
    \begin{tabular}{@{}llccccc@{}}
        \toprule
        Type & Tracker & Speed & MACs & AUC & P & SA \\
        & & (fps$\uparrow$) & (G$\downarrow$) & & & \\
        \midrule
        \multirow{7}{*}{Local}
        & OSTrack~\cite{ye2022ostrack} & 137 & \textbf{29.1} & 53.7 & 73.9 & 54.7 \\
        & ROMTrack~\cite{cai2023romtrack} & 8 & 34.5 & 54.7 & 74.5 & 55.7 \\
        & ZoomTrack~\cite{kou2023zoomtrack} & \textbf{154} & \textbf{29.1} & 58.4 & 81.2 & 59.4 \\
        & DropTrack~\cite{wu2023dropmae} & 151 & \textbf{29.1} & 59.2 & 82.2 & 60.2 \\
        & FocusTrack (SRA)~\cite{wang2025focustrack} & 143 & \textbf{29.1} & 62.3 & 84.1 & 63.2 \\
        & FocusTrack~\cite{wang2025focustrack} & 44 & 30.1 & 62.8 & 86.2 & 63.9 \\
        & CMRTrack (ours) & 62 & 29.2 & \textbf{67.3} & 89.9 & \textbf{68.5} \\
        \midrule
        Global
        & SiamDT~\cite{huang2024antiuav410} & 8 & 225.3 & 66.8 & \textbf{90.0} & 68.2 \\
        \bottomrule
    \end{tabular}%
    }
\end{table}

As shown in Table~\ref{tab:efficiency_comparison}, CMRTrack achieves the best AUC and SA among all compared trackers while maintaining real-time tracking speed. Compared with the OSTrack baseline, CMRTrack improves AUC from 53.7 to 67.3 and SA from 54.7 to 68.5, with only a small increase in computational complexity from 29.1G to 29.2G MACs. Although the practical tracking speed decreases due to the additional motion branch, historical search handling, and reliability-aware score fusion, CMRTrack still runs at 62 fps, satisfying real-time tracking requirements.

Compared with FocusTrack, CMRTrack improves AUC by 4.5 percentage points and SA by 4.6 percentage points, while requiring slightly fewer MACs, 29.2G compared with 30.1G. It is also faster than the full FocusTrack model, 62 fps versus 44 fps. Compared with the global tracker SiamDT, CMRTrack achieves higher AUC and SA with substantially lower computational complexity, requiring only 29.2G MACs compared with 225.3G MACs. These results demonstrate that CMRTrack achieves a favorable balance between tracking accuracy, computational complexity, and real-time efficiency.

\section{CONCLUSION}

This article presents CMRTrack, a counterfactual motion reliability learning framework for robust infrared UAV tracking. CMRTrack extracts temporal cues from adjacent search regions and regularizes motion learning with a counterfactual target-erased history, enabling the tracker to suppress background-induced pseudo motion and emphasize target-consistent temporal evidence. The learned motion evidence is incorporated through motion-guided token modulation and reliability-aware score fusion to improve feature representation and response prediction.
Extensive experiments on challenging Anti-UAV benchmarks show that CMRTrack outperforms representative state-of-the-art trackers while maintaining real-time efficiency. Ablation studies, attribute-based evaluation, and qualitative visualization further demonstrate its robustness under fast motion, scale variation, thermal crossover, out-of-view, and dynamic background clutter. Future work will explore richer temporal priors and multi-frame motion modeling for more complex low-altitude surveillance scenarios.
\vspace*{-.5pc}

\bibliographystyle{IEEEtran}
\bibliography{refs}

\end{document}